\documentclass[letterpaper, 10 pt, journal, twoside]{IEEEtran}

\IEEEoverridecommandlockouts                              

\usepackage[pdftex]{graphicx}
\usepackage{subcaption}
\usepackage{hyperref}

\usepackage{diagbox}

\title{Measuring Object Rotation via Visuo-Tactile Segmentation of Grasping Region
}

\author{Julio Castaño-Amorós$^{1}$ and Pablo Gil$^{2}$

\thanks{Manuscript received: February, 22, 2023; Revised April, 12, 2023; Accepted June, 1, 2023.}

\thanks{This paper was recommended for publication by Editor M. Vincze upon evaluation of the Associate Editor and Reviewers' comments.}

\thanks{This work was supported by the Ministry of Science and Innovation of the Spanish Government through the research project  PID2021-122685OB-I00 and by the University of Alicante through the grant UAFPU21-26.
}
\thanks{$^{1}$Julio Castaño-Amorós is with the AUROVA Lab, Computer Science Research Institute, University of Alicante, Alicante, Spain.
        {\tt\footnotesize julio.ca@ua.es}}%
\thanks{$^{2}$Pablo Gil is with the AUROVA Lab, Computer Science Research Institute and with Department of Physics, Systems Engineering, and Signal Theory, University of Alicante, Alicante, Spain.
        {\tt\footnotesize pablo.gil@ua.es}}%

\thanks{Digital Object Identifier (DOI): see top of this page.}
}

\begin{document}

\maketitle

\begin{abstract}

When carrying out robotic manipulation tasks, objects occasionally fall as a result of the rotation caused by slippage. This can be prevented by obtaining tactile information that provides better knowledge on the physical properties of the grasping. In this paper, we estimate the rotation angle of a grasped object when slippage occurs. We implement a system made up of a neural network with which to segment the contact region and an algorithm with which to estimate the rotated angle of that region. 
This method is applied to DIGIT tactile sensors. Our system has additionally been trained and tested with our publicly available dataset which is, to the best of our knowledge, the first dataset related to tactile segmentation from non-synthetic images to appear in the literature, and with which we have attained results of 95\% and 90\% as regards Dice and IoU metrics in the worst scenario. Moreover, we have obtained a maximum error of $\approx 3$º when testing with objects not previously seen by our system in 45 different lifts. This, therefore, proved that our approach is able to detect the slippage movement, thus providing a possible reaction that will prevent the object from falling. 

\end{abstract}


\section{INTRODUCTION}

\IEEEPARstart{T}{he} methods employed to carry out robotic manipulation tasks, which are based on 2D/3D vision techniques \cite{du}, generally take into account only the geometric properties of objects in the scene.
The physical properties of the objects, such as their mass distribution (variable or otherwise), the center of gravity, or friction are not contemplated, signifying that the object could fall if it is not correctly grasped. In contrast, the use of 
tactile sensors 
makes it possible to measure and react to these physical properties in order to achieve stable grasping \cite{luo}.

Several technologies with which to attain a sense of touch and thus find solutions to complex problems such as slip detection have been designed. The most common are electrical-based tactile sensors, such as capacitive, piezoresistive or optical-based tactile sensors \cite{hardwaretactil}, on which our work is focused.

The most typical event to cause an object to fall from human or robotic hands 
during a manipulation task is probably slippage. 
In order to achieve a stable grasping while a robotic hand is manipulating an object, it is necessary to obtain and interpret information concerning the contact between the sensor surface and the grasped object. Tactile sensors can, therefore, be used to sense the slippage \cite{slippage_review} and the data obtained can then be used to control the movement of the robotic hand.
For example, tactile readings have previously been used: to classify the grasping in slip or non-slip states, as shown in  \cite{slippage_piezoresistive}, to classify the kind of movement (rotation and translation) that takes place when a slippage occurs \cite{slippage_barometric}, 
or to measure displacement by using visual markers on tactile optical sensors \cite{slippage_gelsight}.

In this paper, we present an algorithm with which to estimate the rotation angle of an object that is being manipulated when slippage occurs. This method is applied to optical DIGIT sensors \cite{digit} whose tactile reading is simply an RGB image without visual markers, depth information or force values.
Our main contribution is threefold:
    \begin{itemize}
        \item The implementation of a two-stage method based on deep learning and traditional computer vision algorithms in order to segment the contact region and estimate the rotational slippage angle for the task of grasping and lifting an object. 
        \item The generation and public sharing of our tactile segmentation dataset. 
         This is, to the best of our knowledge, the first dataset related to tactile segmentation in literature containing data from real DIGIT sensors, whose objective is to reduce the manual labeling process and encourage the use of learning techniques in order to expand these data to other units of DIGIT sensors. 
        \item Extensive and rigorous experimentation and the comparison of different segmentation neural networks and computer vision algorithms used for the task of tactile segmentation. We also compare our proposal with other state-of-the-art methods for the 
        estimation of the angle of rotational slippage.
    \end{itemize}

This paper is organized as follows: Section \ref{related_works} provides a brief introduction to related works regarding contact segmentation and angle estimation for slippage detection, while Section \ref{sec:methods} provides a detailed description of our proposed method and how it differs from the other approaches. Section \ref{section:experimentation} shows two comparisons: one of the different variants of the two stages of our method, and the other of our final proposal and other state-of-the-art methods in literature. Finally, the results obtained are summarized and discussed in Section \ref{sec:conclusions}, as are the limitations of this work.

\section{RELATED WORK}
 \label{related_works}
\subsection{Estimating the Contact Region} 

Estimating the contact region 
between the robot’s fingertips and the object being grasped is crucial to the performance of any manipulation task. 
For example, the objective of the work presented in \cite{kolamuri} and \cite{pytouch} was to estimate the contact region 
by subtracting contact and non-contact tactile images. 
In \cite{ito} and \cite{kakani}, the authors used vision-based tactile sensors with markers to estimate the contact region. 
In the first case, it was obtained by detecting and grouping the moving markers, while in the second it was estimated through the use of a Gaussian regression model. 
Other authors, in \cite{suresh2022}, used synthetic tactile images obtained from simulation to train a Residual NN model of contact recognition for 3D reconstruction of object surfaces. Following a similar line, \cite{gelslim} and \cite{gelsight} solved the contact estimation task for the surface reconstruction between the object and the fingertips implementing photometric algorithms.

 Finally, the approach that inspired our work consists of segmenting the contact region with neural networks. In \cite{bauza}, the authors used a vanilla Convolutional Neural Network (CNN) to generate heightmaps from tactile images with the aim of reconstructing the local contact surface. In \cite{lepora} they used Generative Adversarial Networks (GANs) to segment the region of contact in order to track the contour of the object by applying Reinforcement Learning (RL) techniques.

In contrast, in this work, we use tactile segmentation to calculate the rotation angle of an object when slippage occurs during manipulation tasks. Although our work is inspired by the aforementioned works, the main differences lie in the fact that we use DIGIT sensors without markers \cite{ito}, \cite{kakani}, which do not produce depth information \cite{gelslim}, \cite{gelsight}, and state-of-the-art segmentation neural networks, which are more robust than subtracting operations \cite{kolamuri}, \cite{pytouch} and vanilla CNN \cite{bauza}, and whose training is more stable when compared to the training of GANs \cite{lepora}. Moreover, in this paper, our methods are trained in order to segment several contact geometries of real household objects, while in \cite{bauza} the authors trained their CNNs using basic 3D printed geometries, and in \cite{lepora} they trained an RL agent to follow contours and surfaces by segmenting edges.

\subsection{Slip Detection and Estimation of the
Object Rotation}

Slippage is a common physical event that occurs during object manipulation, and attempts to solve it have been made for several years.
For example, in \cite{riai} the authors implemented traditional image preprocessing techniques in order to detect slippage from tactile images. 
In \cite{zhang}, Long Short-Term Memory (LSTM) neural networks were 
trained to identify translation, rotation or rolling slip movements from tactile images. 
In \cite{brayan}, the authors combined CNNs and Recurrent Neural Networks (RNNs) so as to classify 
slippage in translation, clockwise and counterclockwise rotation from sequential pressure values. A more advanced approach is to quantify the translation or rotation. In this line, \cite{kolamuri} used vision-based tactile sensors with markers to calculate the rotation angle from the estimation of the rotation center by employing a least square algorithm. 
Another approach consists of exchanging the vision-based sensors for force/torque sensors in order to obtain the rotation angle in a pivoting task as occurs in \cite{toskov}, in which the authors trained RNNs to calculate the pivoting angle from sequential force values. 

The work presented in this proposal was inspired by the methods that characterize and quantify the rotational slippage 
in order to evaluate the performance of different algorithms, which can be combined in the pipeline of our method, for the estimation of 
the rotation angle from the 
segmented region of predicted contact. We additionally compare our two-stage method with an end-to-end method for this same task.

\section{METHODS}
\label{sec:methods}

\subsection{Our Method for the Estimation of the Slippage Angle}

\begin{figure}[htbp]
     \centering
         \centering
         \includegraphics[width=0.4\textwidth, height=11.5cm]{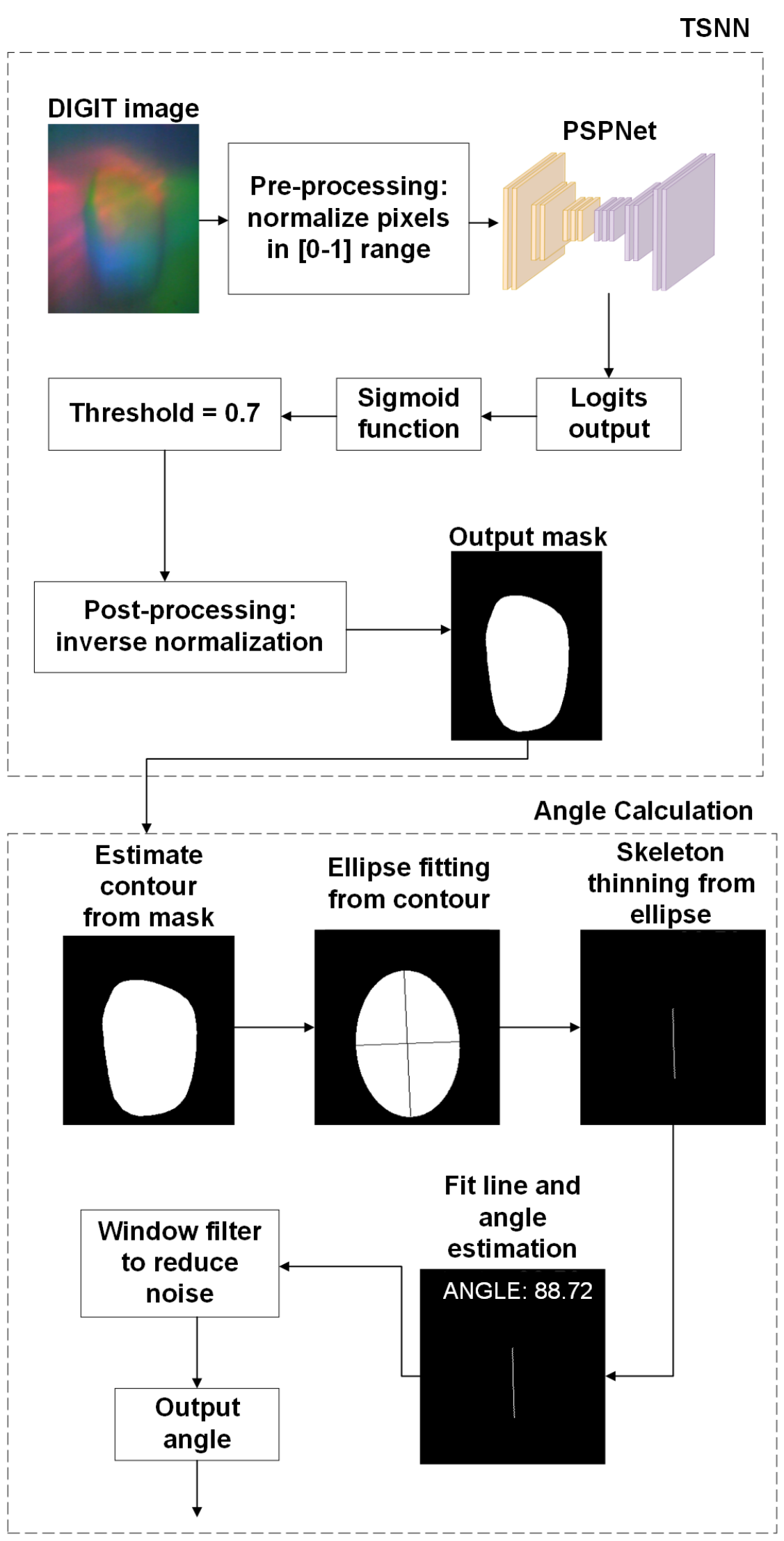}
     \caption{Scheme 
     of our system combining both stages}
     \label{fig:method}
\end{figure}

We propose a two-stage method with which to estimate the angle of rotation caused by slippage during a robotic task concerning the grasping and lifting of an object. Figure \ref{fig:method} shows a scheme that defines the different parts of the two stages of our method. Besides, other algorithms are described to define other approaches to carry out a comparative and ablation study for the validation of our method.

The first stage, denominated as Tactile Segmentation Neural Network (TSNN), receives a raw RGB image from a DIGIT sensor as input, which is normalized by scaling the pixel values in a range of between 0 and 1. This normalized image is sent as input to a PSPNet model \cite{pspnet} that was trained to segment the contact region of the tactile image. A sigmoid activation function is applied so as to transform the output of the PSPNet model, known as logits, into pixel values in a range of [0-1]. Note that this is important in order to later use an empirically obtained restrictive threshold of 0.7 to classify the pixel values from the [0-1] range in two classes: background (0-pixel value) and contact region (1-pixel value). Finally, a post-processing step is carried out, which consists in applying the inverse normalization of the pre-processing step in order to transform the pixel values into a [0-255] range so as to generate the final output mask.

The output mask from the previous stage is used as input in the second stage to obtain the predominant contour present in the image. An ellipse-fitting \cite{ellipse} algorithm is then used to reduce the noise of the segmented output mask, and a Skeleton Thinning \cite{skeleton} algorithm is later applied in order to obtain the main axis of the output mask. This axis can be used to estimate the angle of rotation by fitting a line with a least square process and calculating the arctan function. The angle is processed by a window filter that averages the last $n$ angles in order to produce the mean angle of rotation. This filter helps reduce the noise from the camera's signal, which affects the segmented region by changing its size and shape.

\subsection{Variants of our Method for a Comparative and Ablation Study}

\label{sec:variants_methods}

As stated previously, our method is made up of two stages: one for the contact region segmentation and another for the estimation of the slippage angle. Our TSNN is based on PSPNet architecture, which extracts 
 a feature map from the last layer of a CNN in order to later process it with a Pyramid Pooling module so as to learn features on different scales, and finally concatenate the original and the processed feature maps in order to generate the output mask. Furthermore, our angle estimation method is based principally on the Skeleton Thinning algorithm, which narrows the contact region into a line by blackening white pixels depending on the neighborhood.

 In order to carry out comparative experimentation, two segmentation NNs (DeepLabV3+ \cite{deeplabv3+} and Unet++ \cite{unet++}) adapted to our task and two angle estimators (Principal Component Analysis (PCA) \cite{pca} and Ellipse Fitting \cite{ellipse}) were considered.
 
On the one hand, DeepLabV3+ and Unet++ architectures differ from the PSPNet architecture since they are based on the encoder-decoder scheme. Moreover, while Unet++ is known for re-designing the skip connections in Unet so as to reduce the semantic gap in order to capture fine-grained details, DeepLabV3+ is characterized by the use of a combination of atrous or dilated convolutions and depthwise separable convolutions, which reduce the computational complexity while maintaining the performance. On the other hand, PCA and Ellipse Fitting differ from Skeleton Thinning since they estimate the angle of slippage by calculating the eigenvalues and eigenvectors. In PCA, the covariance matrix is used to calculate the eigenvalues and eigenvectors in order to identify the principal components of the data. In Ellipse Fitting, the LIN algorithm \cite{ellipse} is applied to estimate the curve that best fits a set of input points (in our case, the contour points of the segmented region) and its output is the eigenvectors from which the angle is calculated.

 The following section, therefore, shows an experimental comparison of our proposal (PSPNet-Skeleton) versus the other eight 
 possible combinations of segmentation 
 (DeepLabV3+, UNet++, our TSNN based on PSPNet) and angle estimators (PCA, Ellipse Fitting, Skeleton Thinning) in the comparative and ablation study.

\section{EXPERIMENTATION AND RESULTS}
\label{section:experimentation}

\subsection{Data Acquisition and Dataset Generation}
\begin{figure}[htbp]
     \centering
         \centering
         \includegraphics[scale=0.85]{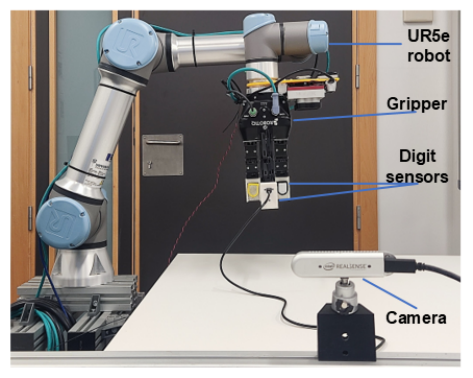}
     \caption{Setup made up of an external camera, a UR5e robot arm with a 3F Robotiq gripper and DIGIT sensors mounted on the fingertips} 
     \label{fig:setup}
\end{figure}

Although the number of works using visual-based tactile sensors has increased over the last few years, we have been unable to find any dataset related to tactile segmentation containing data from real DIGIT sensors as a basis in our experimentation.
We have chosen to work with real instead of synthetic images \cite{suresh2022} to avoid losing contact features that can be presented in non-rigid objects and affect the grasping.
We have, therefore, generated our own dataset by applying an automatic data recording process with a UR5e robot, a 3F Robotiq gripper, and a visual-based tactile sensor known as DIGIT (see Fig. \ref{fig:setup}). The recording process consisted of capturing tactile images from the sensors while the robot was grasping an object, each time with a different pose.
The grasping force was also varied (range $\approx$ [0.2-1.8N]) depending on the mass, rigidity, type of surface and geometry of the grasping area of the object.
Only two fingers were used to grasp the objects in order to test our system in extremely unstable conditions.

\begin{figure}[htbp]
     \centering
         \centering
         \includegraphics[width=0.4\textwidth, height=5cm]{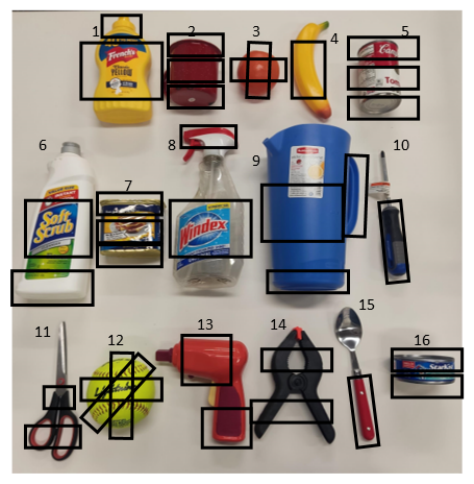}
     \caption{Objects from YCB dataset used to generate our tactile segmentation dataset. \href{https://github.com/AUROVA-LAB/aurova_grasping/tree/main/Tactile_sensing/Digit_sensor/Tactile_segmentation/dataset}{Link to download dataset}}
     \label{fig:tactile_segmentation_dataset}
\end{figure}

Our tactile segmentation dataset was created by using 16 objects (see Fig. \ref{fig:tactile_segmentation_dataset}), and between 200 and 250 tactile images were then captured per object. The objects belong to the YCB dataset \cite{ycb_dataset} and contain different features related to touch sensing, such as texture, rigidity, size, mass, friction, or shape, thus, allowing us to contribute to the repeatability of the experimentation and the benchmarking of tactile manipulation.

We manually defined 15 poses (a combination of 3 positions and 5 orientations) per object, and ensure that our DIGIT sensor was capturing different geometries from the local contact with the object. This number of poses per object was, therefore, sufficient to form our dataset owing to the uniformity and symmetry of the surfaces of the objects. The variability of objects and poses thus  
enabled the neural networks trained with our dataset to achieve high generalization capabilities. The black rectangles in Fig. \ref{fig:tactile_segmentation_dataset} represent the grasping areas established in order to record our dataset so as to capture a large variety of tactile images with different contact geometries. The bigger boxes represent a higher percentage of grasping actions in this area.

In this work, we have formulated the tactile segmentation task as a supervised learning problem, signifying that it was necessary to label our data. This was done by using the LabelMe tool \cite{labelme} to label the contact regions on the tactile samples. The labeling process was carried out manually by observing the contact images with respect to a non-contact image reference. 

The initial tactile segmentation dataset, which contains a total of 3675 images, was then used to generate three sets ($D_{s1}$, $D_{s2}$, and $D_{s3}$) by applying the 3-fold cross-validation technique. This was done in order to carry out an optimal evaluation of our models. We randomly split each set by following a distribution of 70\% of the objects (11 objects) for training, 20\% (3) for validation, and 10\% (2) for testing. Table \ref{tab:k-fold_cross_validation} shows the corresponding identifiers of the objects shown in Fig. \ref{fig:tactile_segmentation_dataset} for each set.

\begin{table}[htpb]
\centering
\caption{3-fold cross-validation technique and mapping between our tactile segmentation dataset and object's identifiers}
\label{tab:k-fold_cross_validation}
 \resizebox{0.48\textwidth}{!}{  
 \renewcommand{\arraystretch}{1.2}
\begin{tabular}{|c|c|c|c|}
\hline
                        & $\mathbf{D_{s1}}$                                                                   & $\mathbf{D_{s2}}$                                                                     & $\mathbf{D_{s3}}$                                                                    \\ \hline
\textbf{Train set}      & \begin{tabular}[c]{@{}c@{}}1, 2, 4, 5, 6, 8, 10,\\ 12, 13, 14, 15\end{tabular} & \begin{tabular}[c]{@{}c@{}}1, 2, 3, 4, 6, 7, \\ 10, 11, 12, 13, 16\end{tabular} & \begin{tabular}[c]{@{}c@{}}1, 3, 5, 6, 7, 8, 9, \\ 10, 11, 13, 14\end{tabular} \\ \hline
\textbf{Validation set} & 7, 9, 11                                                                      & 5, 9, 14                                                                        & 4, 15, 16                                                                      \\ \hline
\textbf{Test set}       & 3, 16                                                                         & 8, 15                                                                           & 2, 12                                                                          \\ \hline
\end{tabular}
}
\end{table}

\subsection{
Performance Evaluation of our Proposal vs. other Approaches 
}

The performance of the trained models was evaluated using the $Dice$ and $IoU$ scores, which are described in Eqs. \ref{eq:dice_score} and \ref{eq:iou_score}, respectively.

\begin{equation}
    Dice = \frac{2 * pred * y}{pred + y}
    \label{eq:dice_score}
\end{equation}

\begin{equation}
    IoU = \frac{pred * y}{pred + y - pred * y}
    \label{eq:iou_score}
\end{equation}

where $pred$ are the images containing the predicted contact regions and $y$ are the images with the labeled contact regions. The $Dice$ score is a metric that is similar to the average performance, while the $IoU$ score is more like the worst-case scenario.

The training phase was carried out employing an NVIDIA A100 Tensor Core GPU with 40 GB of RAM, along with
the following optimal hyperparameters obtained from a process of searching 
and tuning: a batch size of 32, a learning rate of 1e-4, the Adam optimizer, the Focal loss, 30 training epochs and a ResNet18 backbone. We also trained with the EfficientNet-B3 backbone, but as ResNet18 achieved higher metric values in all of the training sessions, we have not included the experimental results in our work. ResNet architecture is well-known for
the residual blocks that help reduce the vanishing gradient
problem in deep neural networks. The testing phase was carried out using an NVIDIA GeForce GTX 1650 Ti with 4GB of RAM to perform the comparison regarding real-time capabilities.

Table \ref{tab:unet++_results} shows the results of the testing phase in terms of the $Dice$, $IoU$ and time values 
when our TSNN was compared to the other approaches adapted to the task using 
each set ($D_{s1}$, $D_{s2}$, and $D_{s3}$) of our dataset.

\begin{table}[htpb]
 \caption{Segmentation results for the sets $D_{s1}$, $D_{s2}$, and $D_{s3}$ comparing our 
 TSNN with Unet++ and DeepLabV3+, 
 using the ResNet18 backbone}
\label{tab:unet++_results}
\centering
\renewcommand{\arraystretch}{1.3}
\begin{tabular}{|c|ccc|}
\hline
\multicolumn{4}{|c|}{\textbf{Unet++}}                                                                                 \\ \hline
\textbf{}    & \multicolumn{1}{c|}{\textbf{Dice}}           & \multicolumn{1}{c|}{\textbf{IoU}}            & \textbf{Time(s)}        \\ \hline
$D_{s1}$          & \multicolumn{1}{c|}{0.950 $\pm$ 0.018}          & \multicolumn{1}{c|}{0.906 $\pm$ 0.033}          & 0.008 $\pm$ 0.001            \\ \hline
$D_{s2}$          & \multicolumn{1}{c|}{0.955 $\pm$ 0.008}          & \multicolumn{1}{c|}{0.915 $\pm$ 0.015}          & 0.009 $\pm$ 0.002          \\ \hline
{$\mathbf{D_{s3}}$} & \multicolumn{1}{c|}{\textbf{0.969 $\pm$ 0.006}} & \multicolumn{1}{c|}{\textbf{0.940 $\pm$ 0.011}} & \textbf{0.009 $\pm$ 0.001} \\ \hline
Avg          & \multicolumn{1}{c|}{0.958 $\pm$ 0.011}          & \multicolumn{1}{c|}{0.920 $\pm$ 0.020}          & 0.009 $\pm$ 0.001          \\ \hline
\multicolumn{4}{|c|}{\textbf{DeepLabV3+}} 
       \\ \hline
\textbf{}    & \multicolumn{1}{c|}{\textbf{Dice}}           & \multicolumn{1}{c|}{\textbf{IoU}}            & \textbf{Time(s)} \\ \hline
\multicolumn{1}{|c|}{$D_{s1}$}          & \multicolumn{1}{c|}{0.947 $\pm$ 0.020}          & \multicolumn{1}{c|}{0.899 $\pm$ 0.036}          & 0.007 $\pm$ 0.002          \\ \hline
\multicolumn{1}{|c|}{$D_{s2}$}          & \multicolumn{1}{c|}{0.951 $\pm$ 0.012}          & \multicolumn{1}{c|}{0.906 $\pm$ 0.021}          & 0.006 $\pm$ 0.002          \\ \hline
\multicolumn{1}{|c|}{$\mathbf{D_{s3}}$} & \multicolumn{1}{c|}{\textbf{0.969 $\pm$ 0.007}} & \multicolumn{1}{c|}{\textbf{0.937 $\pm$ 0.013}} & \textbf{0.006 $\pm$ 0.002} \\ \hline
\multicolumn{1}{|c|}{Avg}          & \multicolumn{1}{c|}{0.956 $\pm$ 0.013}          & \multicolumn{1}{c|}{0.914 $\pm$ 0.023}          & 0.006 $\pm$ 0.002          \\ \hline
\multicolumn{4}{|c|}{\textbf{TSNN}}                                   
  \\ \hline
  \textbf{}    & \multicolumn{1}{c|}{\textbf{Dice}}           & \multicolumn{1}{c|}{\textbf{IoU}}            & \textbf{Time(s)} \\ \hline
\multicolumn{1}{|c|}{$D_{s1}$}          & \multicolumn{1}{c|}{0.937 $\pm$ 0.026}          & \multicolumn{1}{c|}{0.883 $\pm$ 0.045}          & 0.007 $\pm$ 0.003          \\ \hline
\multicolumn{1}{|c|}{$D_{s2}$}          & \multicolumn{1}{c|}{0.953 $\pm$ 0.009}          & \multicolumn{1}{c|}{0.910 $\pm$ 0.017}          & 0.006 $\pm$ 0.003          \\ \hline
\multicolumn{1}{|c|}{$\mathbf{D_{s3}}$} & \multicolumn{1}{c|}{\textbf{0.963 $\pm$ 0.008}} & \multicolumn{1}{c|}{\textbf{0.929 $\pm$ 0.014}} & \textbf{0.006 $\pm$ 0.001} \\ \hline
\multicolumn{1}{|c|}{Avg}          & \multicolumn{1}{c|}{0.951 $\pm$ 0.014}          & \multicolumn{1}{c|}{0.907 $\pm$ 0.025}          & 0.006 $\pm$ 0.002          \\ \hline
\end{tabular}
\end{table}

\begin{figure}[htbp]
     \centering
         \centering
         \includegraphics[width=0.47\textwidth]{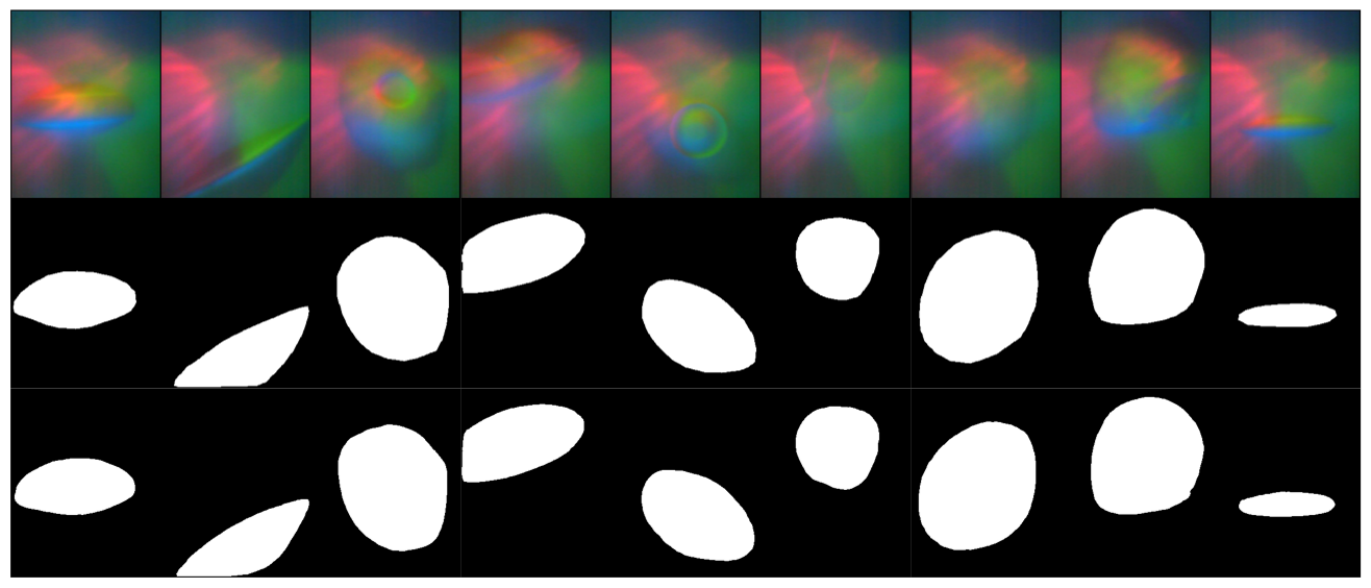}
     \caption{Examples of tactile segmentation made by our TSNN. 
     The first row corresponds to the raw tactile images, 
     the second row is the ground truth segmented images and the third row corresponds to the predicted contact region} 
     \label{fig:examples_tactile_segmentation}
\end{figure}

It will be first noted that the $Dice$ scores are higher than the $IoU$ scores in every case. 
It will be also noted that the difference in the average offline inference time is not significant, signifying that the three neural architectures can make predictions in real time. Finally, note that all the approaches, including our TSNN, achieve
higher metric values with the $D_{s3}$ set. This may be for several reasons, such as the fact that the objects in the $D_{s3}$ test set are easier to segment than are those in the other test sets. However, there are no significant differences in the difficulty of segmenting
the objects from each set.
We consequently believe that the $D_{s3}$ set contains a better distribution of contact surfaces that yield higher metric values. We therefore decided that the weights obtained from the training with the $D_{s3}$ set would be used 
for the study and comparison of our TSNN versus the remaining segmentation approaches.
Figure \ref{fig:examples_tactile_segmentation} shows examples of our tactile segmentation with different samples from the test sets.

\begin{figure}[htbp]
     \centering
         \centering
         \includegraphics[scale=0.37]{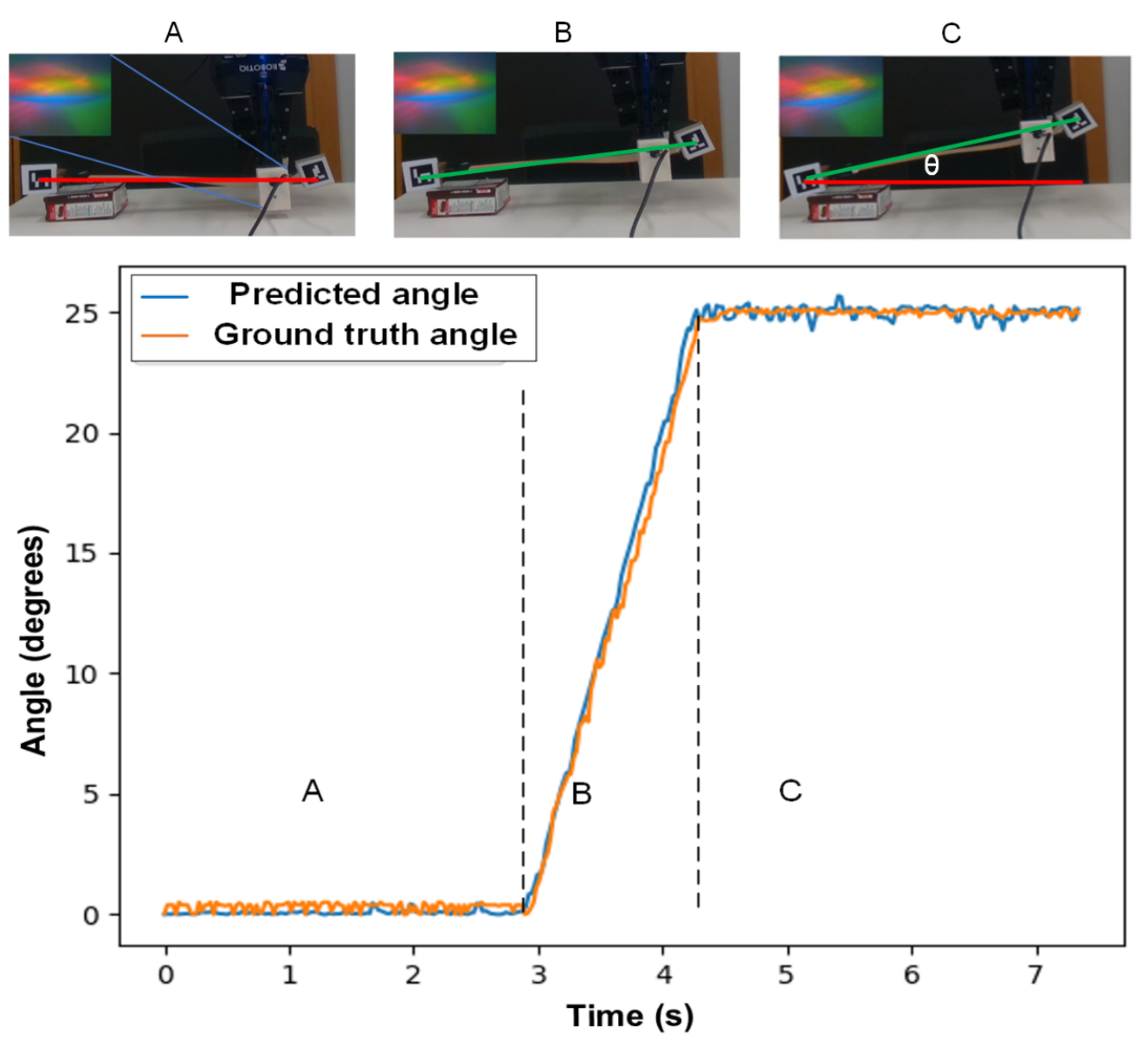}
     \caption{Description of rotation angle calculation during the manipulation task with object 2 from our set of experimentation} 
     \label{fig:describe_angle_rotation_task}
\end{figure}

The performance was evaluated by implementing the task of grasping and lifting an object while our system carried out the tactile segmentation and angle calculation of the segmented contact region. As shown in Fig. \ref{fig:describe_angle_rotation_task}, this task is divided into three parts: the first when the robot grasps the object, the second when the robot lifts it, and the last when the robot holds it on one side. In this work, we focus solely on the second part (part B in Fig. \ref{fig:describe_angle_rotation_task}) because this is where the slippage movement may occur. 

Figure \ref{fig:describe_angle_rotation_task} shows the predicted angle in blue and the ground truth angle in orange. The predicted angle is calculated as being the difference between the current and the initial angle obtained from the methods described in Section \ref{sec:methods}. The ground truth angle $\theta$ is obtained with the formula described in Eq. \ref{eq:arccos_theta}, where $\vec{p}$ is the current vector (green line) and $\vec{q}$ is the initial vector (red line) that links the two ArUco markers.

\begin{equation}
    \theta = acos (\frac{\vec{p} \cdot \vec{q} }{|\vec{p}| \cdot |\vec{q}|})
    \label{eq:arccos_theta}
\end{equation}

The noise from the camera's signal affects the segmented region, thus making the angle calculation noisier. We therefore applied a filtering window that consisted of calculating the mean of $n$ consecutive angles (predicted angle).

\begin{figure}[htbp]
     \centering
         \centering
         \includegraphics[scale=0.85]{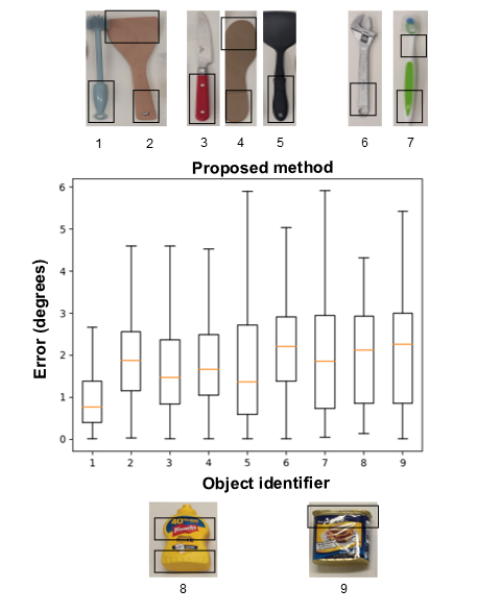}
     \caption{REs for each object using the proposed method (PSPNet and Skeleton Thinning) with a window size of 2 }
     \label{fig:skeleton_best_method}
\end{figure}

We validated our system by randomly selecting two objects from our dataset, concretely, objects 1 and 7 from Fig. \ref{fig:tactile_segmentation_dataset}, and testing it with seven unseen objects (1 to 7 in Fig. \ref{fig:skeleton_best_method}). These nine objects form the dataset ($D_{a}$) for the slip angle experimentation, which differs from the dataset shown in Fig. \ref{fig:tactile_segmentation_dataset} because the objective was to test our proposal with new objects and grasping surfaces.

 The experimentation consisted of making five lifts per object (45 in total) while calculating the \textbf{Rotational Error (RE)} as the absolute difference in degrees between the predicted and ground truth angles. The results have been divided into two parts. 
 
First, a comparative and ablation study was carried out with the proposed method and the other eight possible approaches from Section \ref{sec:variants_methods}. In this study, the \textbf{Mean Absolute Rotational Error (MARE)} was calculated for each of the 9 methods, varying the size of the window filter from 2 to 10 with a step of 2, as can be seen in Table \ref{tab:ablation_study}. In order to compute the MARE, we calculated the mean RE and standard deviation for each of the 45 lifts with each method and window size $n$. We then computed the average of these 45 mean REs and standard deviations in order to obtain the MARE. Note that a 2-size window filter yields the lowest MARE for every method. Increasing the window size does not improve the results because when we increase the window size, the noise is filtered, but the angle magnitude is reduced, thus increasing the error between the predicted and ground truth angle.
 The Skeleton algorithm 
 obtained the lowest MARE with all segmentation methods, no matter which window size, obtaining almost half the error of the other angle approaches. This is because the Skeleton algorithm simplifies the segmented contact region to a single line and does not, therefore, calculate the angle directly from the segmented mask as the PCA and Ellipse approaches do. Small changes in size or translations of the segmented contact region consequently affect the angle calculation to a lesser extent. In summary, the MAREs obtained are similar, although the results obtained with our proposal when combining a TSNN based on PSPNet and Skeleton Thinning are slightly better.

\begin{table*}[htpb]
    \centering
    \caption{Comparison 
    of our proposal, TSNN based on PSPNet and Skeleton Thinning, 
    versus the rest of the approaches 
    }
    \label{tab:ablation_study}
    \renewcommand{\arraystretch}{1.5}
    \begin{tabular}{|l|c|c|c|c|c|}
    \hline
        \backslashbox{\textbf{Method}}{\textbf{Window Size}} & \textbf{2} & \textbf{4} & \textbf{6} & \textbf{8} & \textbf{10} \\ \hline
        DeepLabV3+ and PCA & 3.52º$\pm$1.71º & 3.81º$\pm$1.66º & 4.13º$\pm$1.61º & 4.42º$\pm$1.55º & 4.76º$\pm$1.52º \\ \hline
        DeepLabV3+ and Ellipse & 2.82º$\pm$1.40º & 3.06º$\pm$1.37º & 3.36º$\pm$1.36º & 3.65º$\pm$1.32º & 3.97º$\pm$1.32º \\ \hline
        \textbf{DeepLabV3+ and Skeleton} & \textbf{1.85º$\pm$0.99º} & \textbf{2.02º$\pm$0.96º} & \textbf{2.18º$\pm$0.97º} & \textbf{2.37º$\pm$1.01º} & \textbf{2.62º$\pm$1.04º} \\ \hline
        UNET++ and PCA & 3.48º$\pm$1.62º & 3.78º$\pm$1.59º & 4.05º$\pm$1.56º & 4.33º$\pm$1.50º & 4.66º$\pm$1.47º \\ \hline
        UNET++ and Ellipse & 2.93º$\pm$1.40º & 3.17º$\pm$1.39º & 3.43º$\pm$1.39º & 3.72º$\pm$1.36º & 4.03º$\pm$1.36º \\ \hline
        \textbf{UNET++ and Skeleton} & \textbf{1.90º$\pm$0.90º} & \textbf{2.01º$\pm$0.93º} & \textbf{2.20º$\pm$0.95º} & \textbf{2.44º$\pm$0.99º} & \textbf{2.67º$\pm$1.06º} \\ \hline
        PSPNet and PCA & 3.56º$\pm$1.71º & 3.86º$\pm$1.67º & 4.18º$\pm$1.63º & 4.50º$\pm$1.57º & 4.81º$\pm$1.54º \\ \hline
        PSPNet and Ellipse & 2.90º$\pm$1.39º & 3.14º$\pm$1.37º & 3.42º$\pm$1.37º & 3.72º$\pm$1.35º & 4.02º$\pm$1.37º \\ \hline
        \textbf{Proposed method} & \textbf{1.85º$\pm$0.96º} & \textbf{2.02º$\pm$0.94º} & \textbf{2.19º$\pm$0.97º} & \textbf{2.41º$\pm$1.03º} & \textbf{2.63º$\pm$1.09º} \\ \hline
    \end{tabular}
\end{table*}

 Having proved which method and window size is optimal, we now show the RE for the 5 lifts of each object in the boxplot graphic in Fig. \ref{fig:skeleton_best_method}.
 
Note that the horizontal yellow lines represent the median RE, the box represents the interquartile range and the vertical lines represent the minimum and maximum REs. Overall, our proposal 
achieves a MARE of \textbf{1.85º $\pm$ 0.96º} (see Table \ref{tab:ablation_study}), signifying that our approach estimates the angle of rotational slippage with a maximum MARE of 2.81º in the worst case. Note that the maximum REs are obtained when testing with the objects 5 and 7. This is owing to the geometry of the grasping surface and the segmented region of contact. However, these maximum REs are isolated cases of one of the 5 lifts.

Following the same guidelines described above, the black rectangles in Fig. \ref{fig:skeleton_best_method} represent the different areas of the objects used as the grasping contact origin for the different trials. Note that we set up the contact origins in an attempt to reproduce the way in which a human operator would do so by, for example, avoiding dangerous areas of the objects such as the sharp part of the knife.

\begin{figure}[htbp]
     \centering
     \begin{subfigure}[b]{0.24\textwidth}
         \centering
         \includegraphics[width=\textwidth, height=4cm]{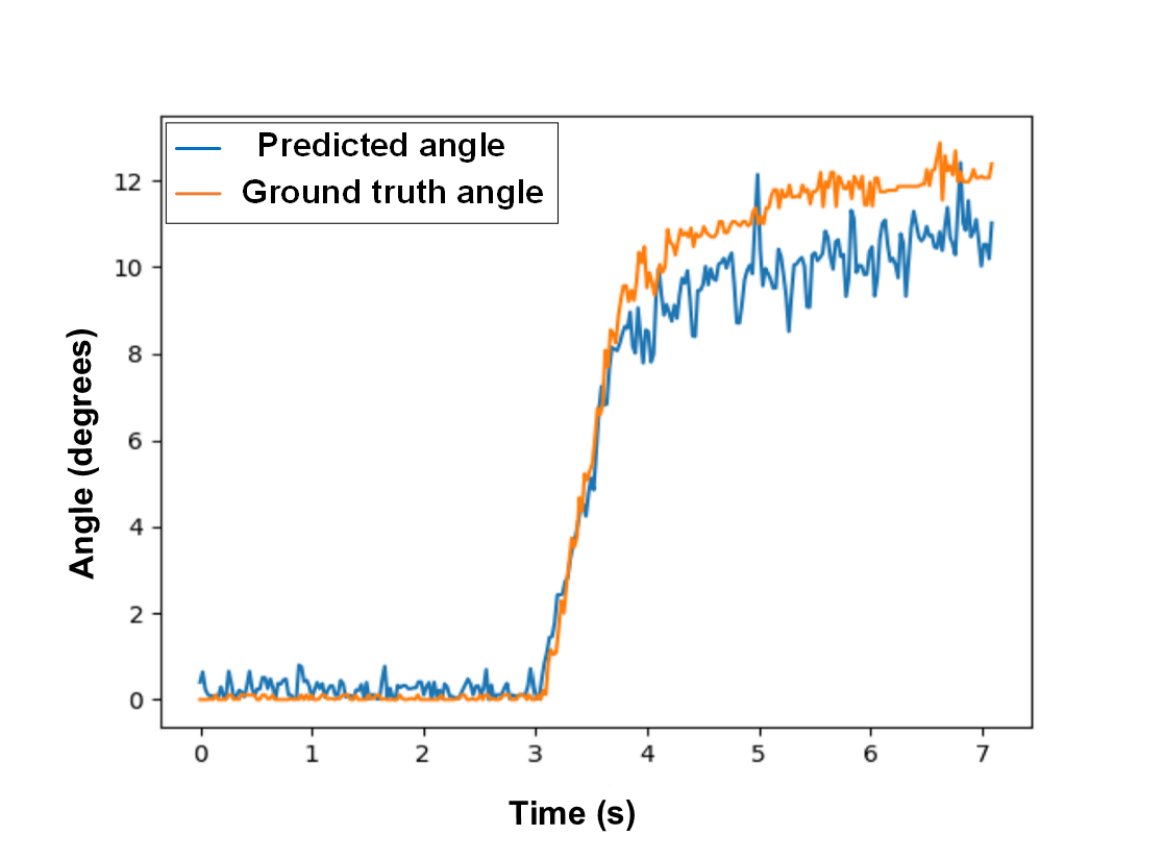}
     \end{subfigure}
     \begin{subfigure}[b]{0.24\textwidth}
         \centering
         \includegraphics[width=\textwidth, height=4cm]{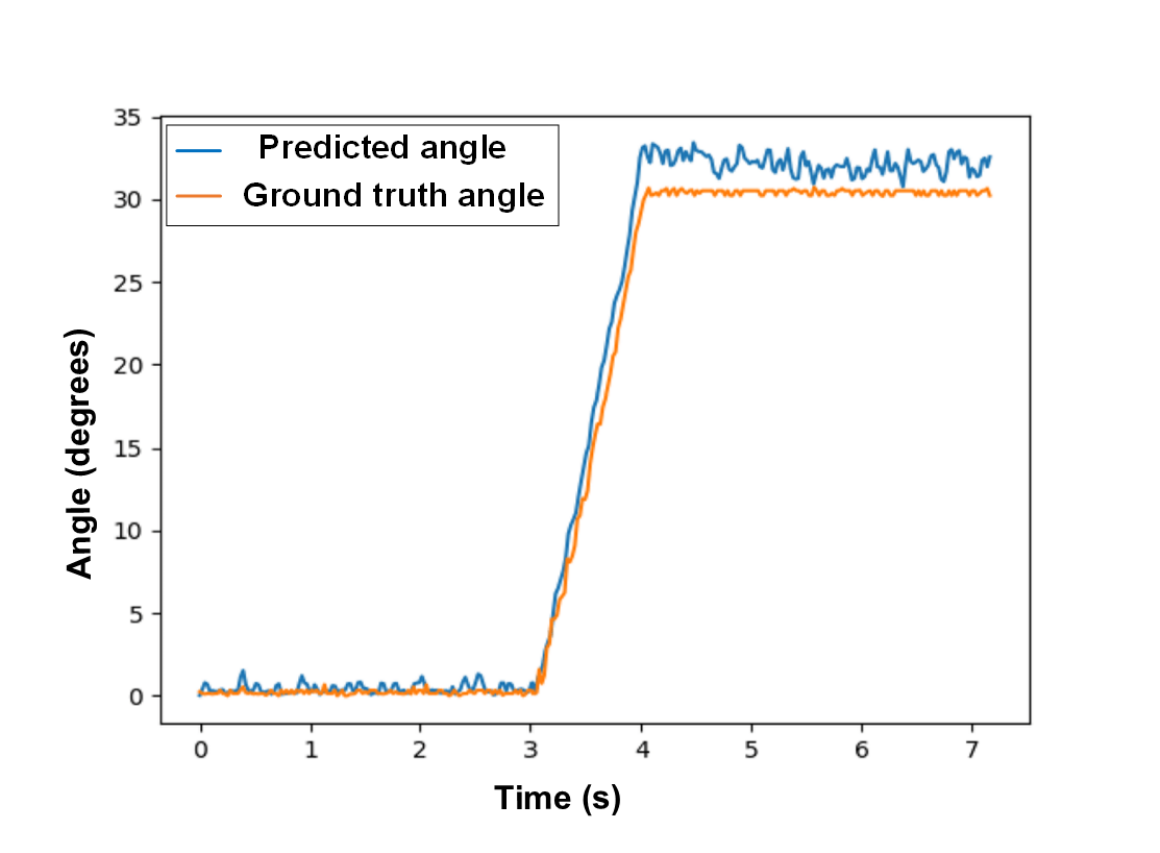}
     \end{subfigure}

    \begin{subfigure}[b]{0.24\textwidth}
         \centering
         \includegraphics[width=\textwidth, height=4cm]{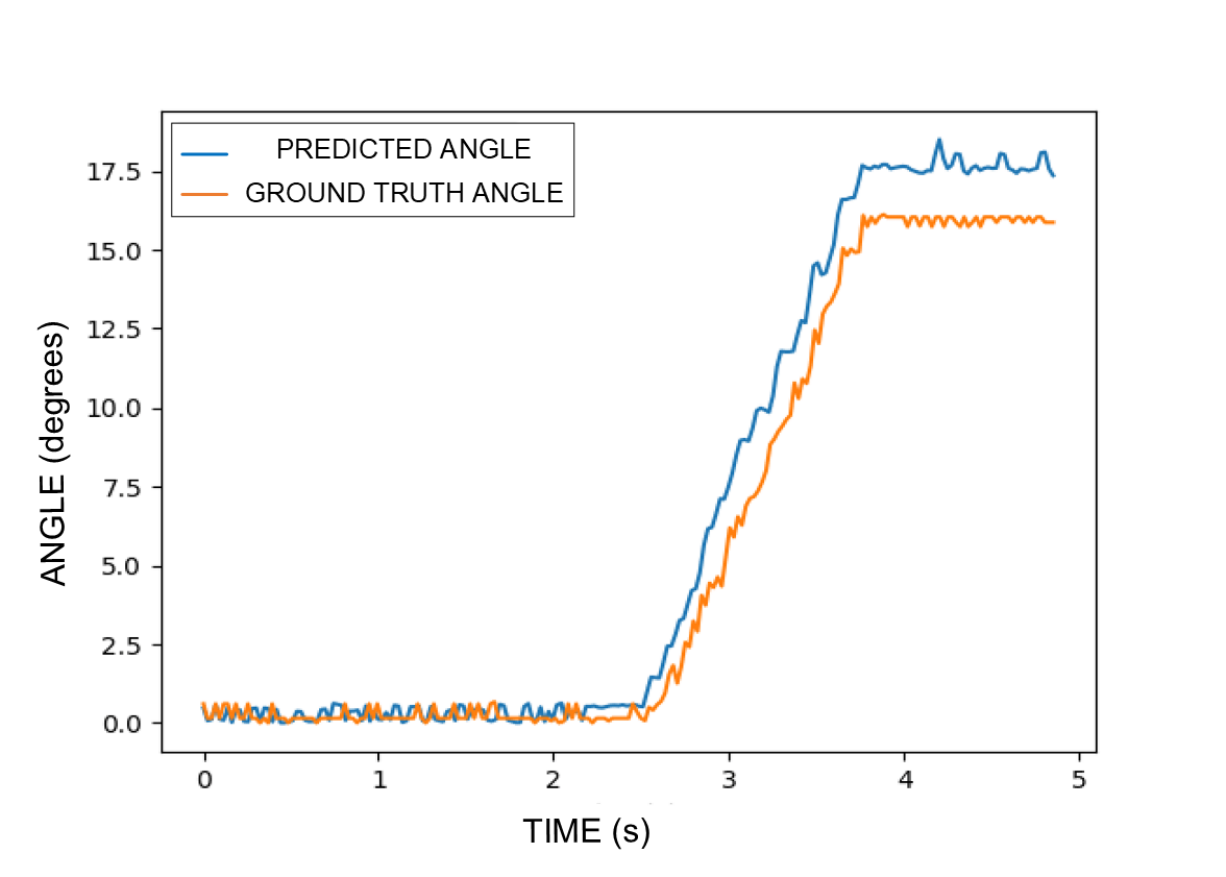}
     \end{subfigure}
     \begin{subfigure}[b]{0.24\textwidth}
         \centering
         \includegraphics[width=\textwidth, height=4cm]{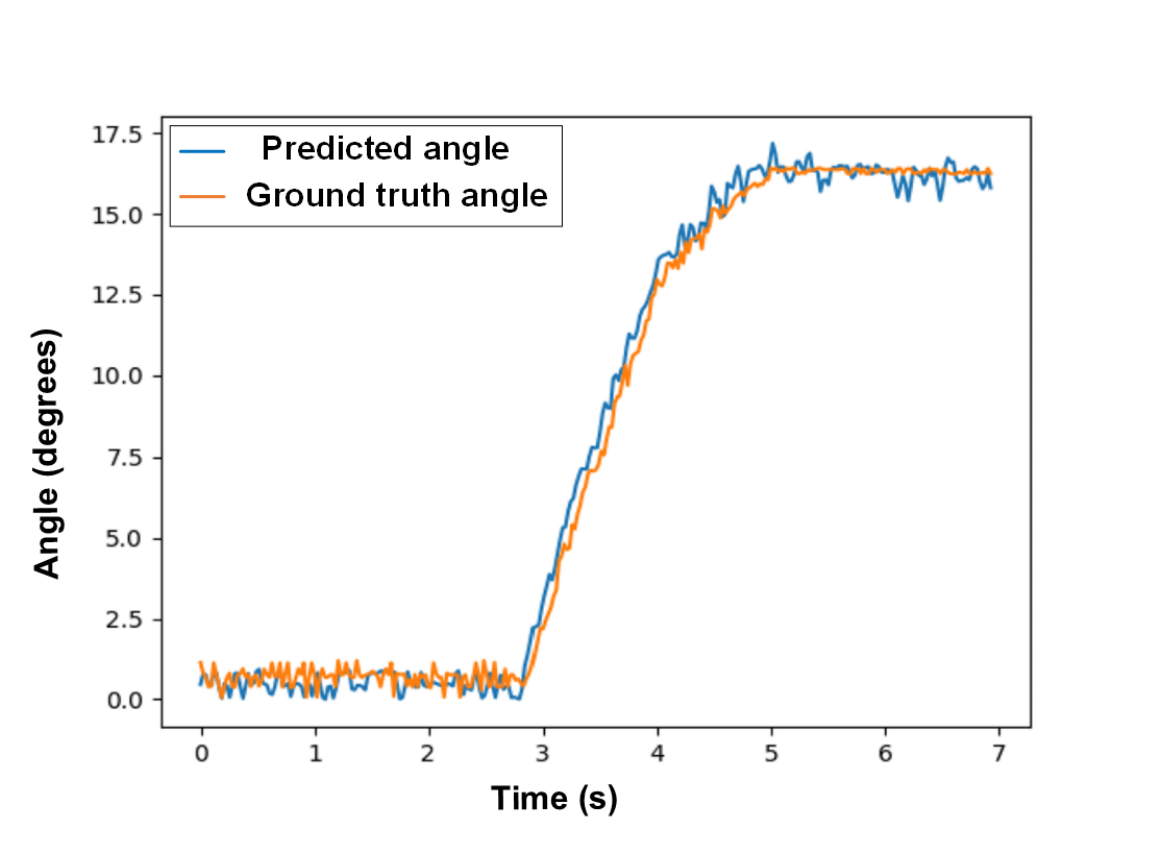}
     \end{subfigure}
     
     \begin{subfigure}[b]{0.24\textwidth}
         \centering
         \includegraphics[width=\textwidth, height=4cm]{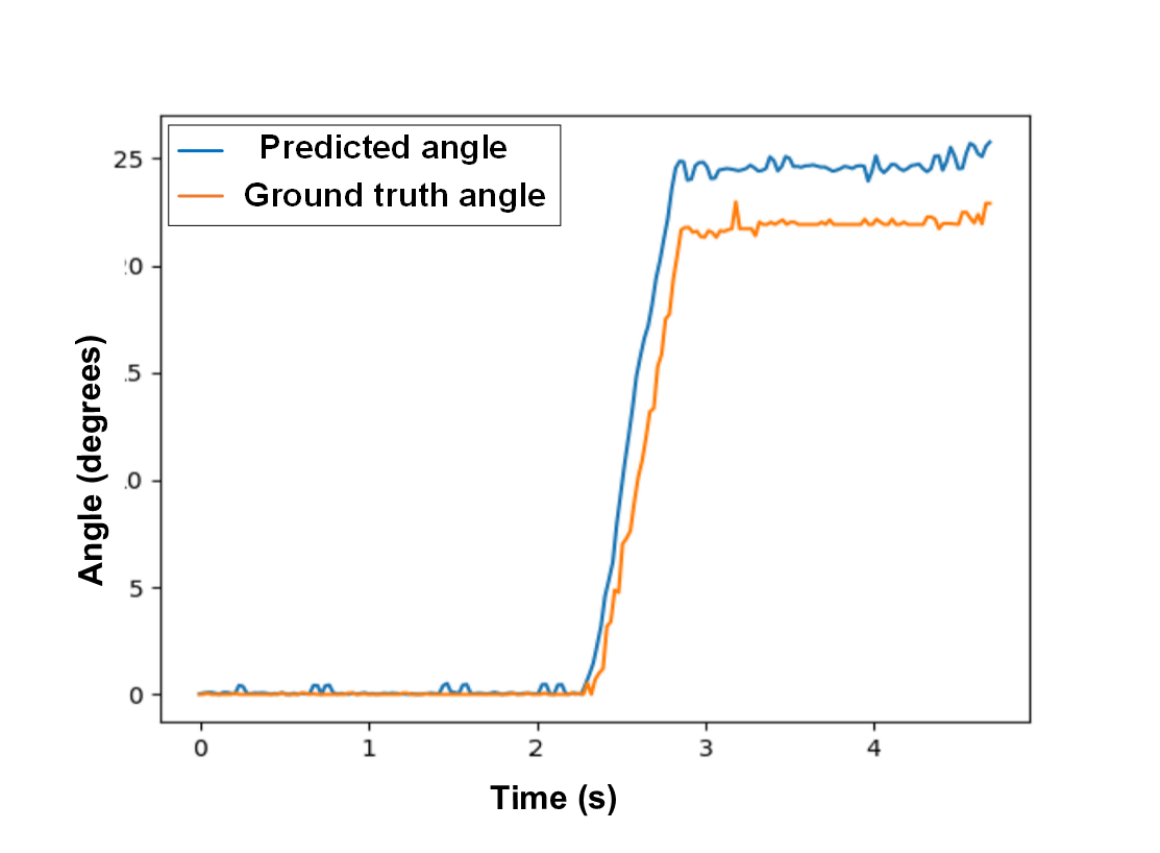}
     \end{subfigure}
     \begin{subfigure}[b]{0.24\textwidth}
         \centering
         \includegraphics[width=\textwidth, height=4cm]{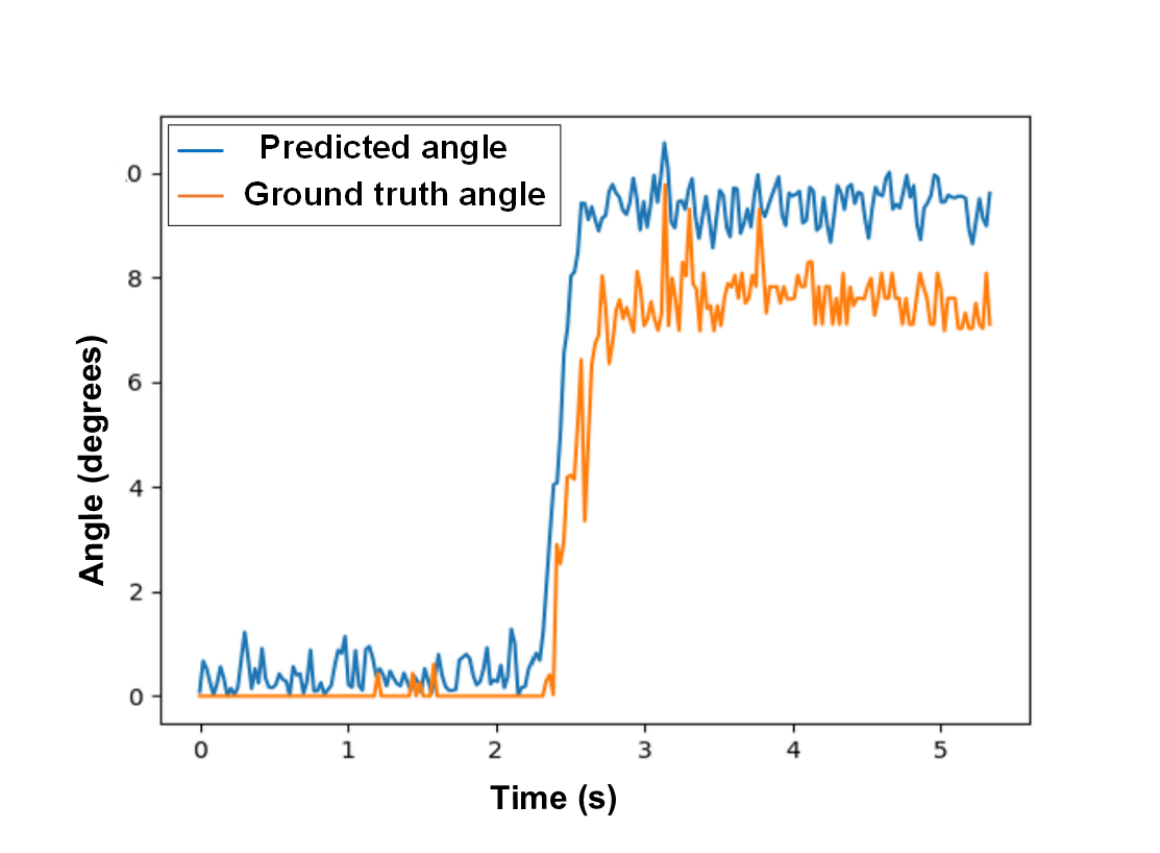}
     \end{subfigure}

     \caption{Examples of REs calculation for slipping during lift task
     with different objects from Fig. \ref{fig:skeleton_best_method}. First row: objects 1, 3. Second row: 4, 5. Third row: objects 6, and 7} 
     \label{fig:examples_angle_calculation}
\end{figure}

Figures \ref{fig:describe_angle_rotation_task} and \ref{fig:examples_angle_calculation} show some examples of RE calculation using our two-stage method 
with different objects. Note that the slippage angle varies in a range between of 0º and 30º for all the lifts because we have limited the angle measurement considering that a robot should react before the slippage angle becomes this value in order to prevent the object from falling.

The error of the measurement depends on many factors, such as the contact shape
, the mass and geometry of the object, etc. The error obviously also depends on the magnitude of the slippage, which is usually higher when the slippage magnitude is higher, but it is not directly related, as can be observed in the case of the objects 5 and 7 in Fig. \ref{fig:examples_angle_calculation}. In this case, object 7 gets a higher error due to a small distance between the grasping pose and the center of mass of the object. 

\subsection{Objective and Subjective Comparison of our proposal vs State-of-the-Art Methods}

As it is complex to find benchmarks for the proposed task because of the variability of data and sensors, we carried out two comparisons. First, a subjective comparison between the proposed method and two other works \cite{kolamuri}, \cite{toskov} was performed. This comparison was subjective owing to the fact that we were unable to reproduce the results obtained in the aforementioned studies because the sensors employed were not available, although the test was carried out with the same or similar objects. Second, we show the results of an objective comparison of our work with another \cite{comparison}. Here, objective signifies that we implemented our own version of the other authors' work in order to reproduce their results and compare them with ours using the exact same objects and data, as their code was not available.

On the one hand,
 \cite{kolamuri} approached the task of slip detection using vision-based tactile sensors containing visual markers and a least-square algorithm, while \cite{toskov} employed force tactile sensors and RNNs to predict the pivoting angle during slippage.
 Although we did not carry out the test with the exact set of objects, it was possible to make a subjective comparison in order prove that competitive results could be attained for this task while using poorer tactile readings, such as only an RGB image. The comparison is shown in the right-hand column of Table \ref{tab:obj_subj_comparison}.

On the other hand, \cite{comparison} presented an NN denominated AngleNet, based on the ConvNext neural architecture, that was used to predict the rotation angle, from DIGIT images, with respect to the vertical axis of tubular objects in Sim-to-Real tasks.
In order to compare AngleNet with our work in the most competitive scenario, we trained it with the 45 lifts from our previous experimentation, and we also tested it with the training data so as to obtain the results in the best possible case. The results obtained are shown in the left-hand column of Table \ref{tab:obj_subj_comparison}. The higher error of AngleNet may be owing to the fact that in an end-to-end approach it is harder to estimate the angle from the tactile reading without knowing where the contact shape is and what it is like.

\begin{table}[htpb]
\centering
\caption{Objective and subjective comparison of the proposed method and other state-of-the-art approaches regarding MARE error and standard deviation}
 \resizebox{0.35\textwidth}{!}{  
\label{tab:obj_subj_comparison}
\renewcommand{\arraystretch}{1}
\begin{tabular}{|c|c|c|}
\hline
                  & \multicolumn{1}{c|}{\textbf{Objective}} & \textbf{Subjective} \\ \hline
\textbf{\cite{kolamuri}} & \multicolumn{1}{c|}{-}                  & 4.39º $\pm$ 0.18º      \\ \hline
\textbf{\cite{toskov}}   & \multicolumn{1}{c|}{-}                  & 3.96º $\pm$ UNK        \\ \hline
\textbf{\cite{comparison}}      & \multicolumn{1}{c|}{3.23º $\pm$ 1.69º}     & -                   \\ \hline
\textbf{Ours}     & \textbf{1.85º 
$\pm$ 0.96º} & \textbf{1.85º $\pm$ 0.96º}                           \\ \hline
\end{tabular}
}
\end{table}

\subsection{Discussion and Limitations of our Method}

\begin{figure}[htbp]
     \centering
     \begin{subfigure}[b]{0.08
     \textwidth}
         \centering
         \includegraphics[height=2.5cm]{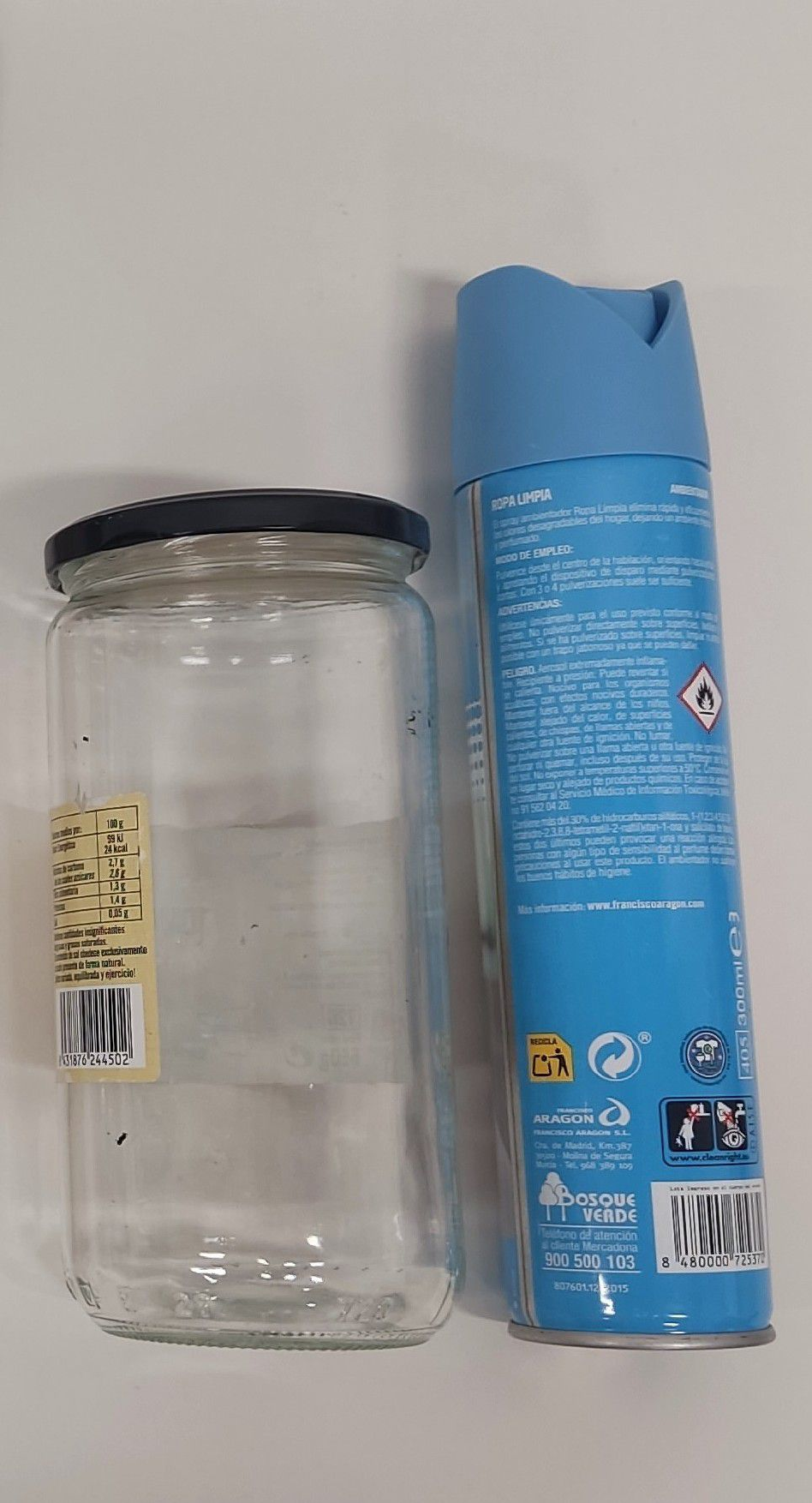}
         \caption{}
     \end{subfigure}
     \begin{subfigure}[b]{0.11\textwidth}
         \centering
         \includegraphics[height=2.5cm, width=2cm]{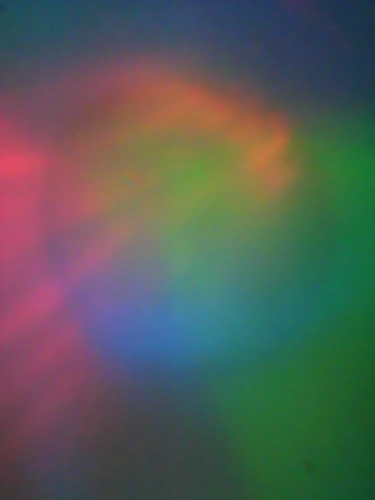}
         \caption{}
     \end{subfigure}
     \begin{subfigure}[b]{0.103\textwidth}
         \centering
         \includegraphics[height=2.5cm, width=2cm]{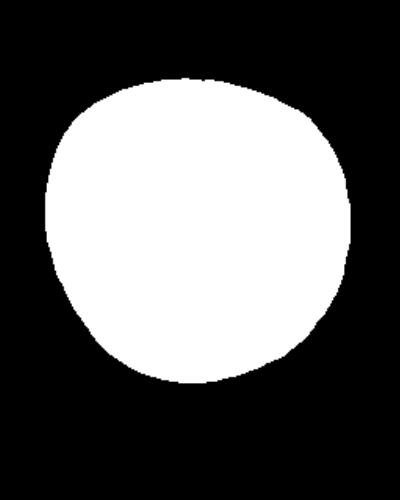}
         \caption{}
     \end{subfigure}
     \begin{subfigure}[b]{0.13\textwidth}
         \centering
         \includegraphics[height=2.5cm, width=2cm]{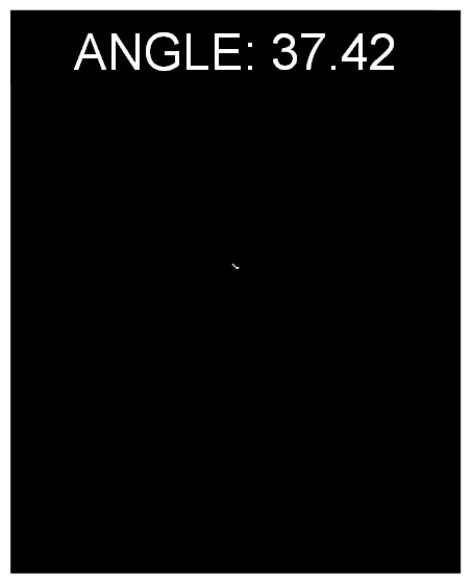}
         \caption{}
     \end{subfigure}
     
     \caption{Example of angle calculation when the segmented contact region is almost circular} 
     \label{fig:limitations1}
\end{figure}

\begin{figure}[htbp]
     \centering

     \begin{subfigure}[b]{0.24\textwidth}
         \centering
         \includegraphics[width=\textwidth, height=4cm]{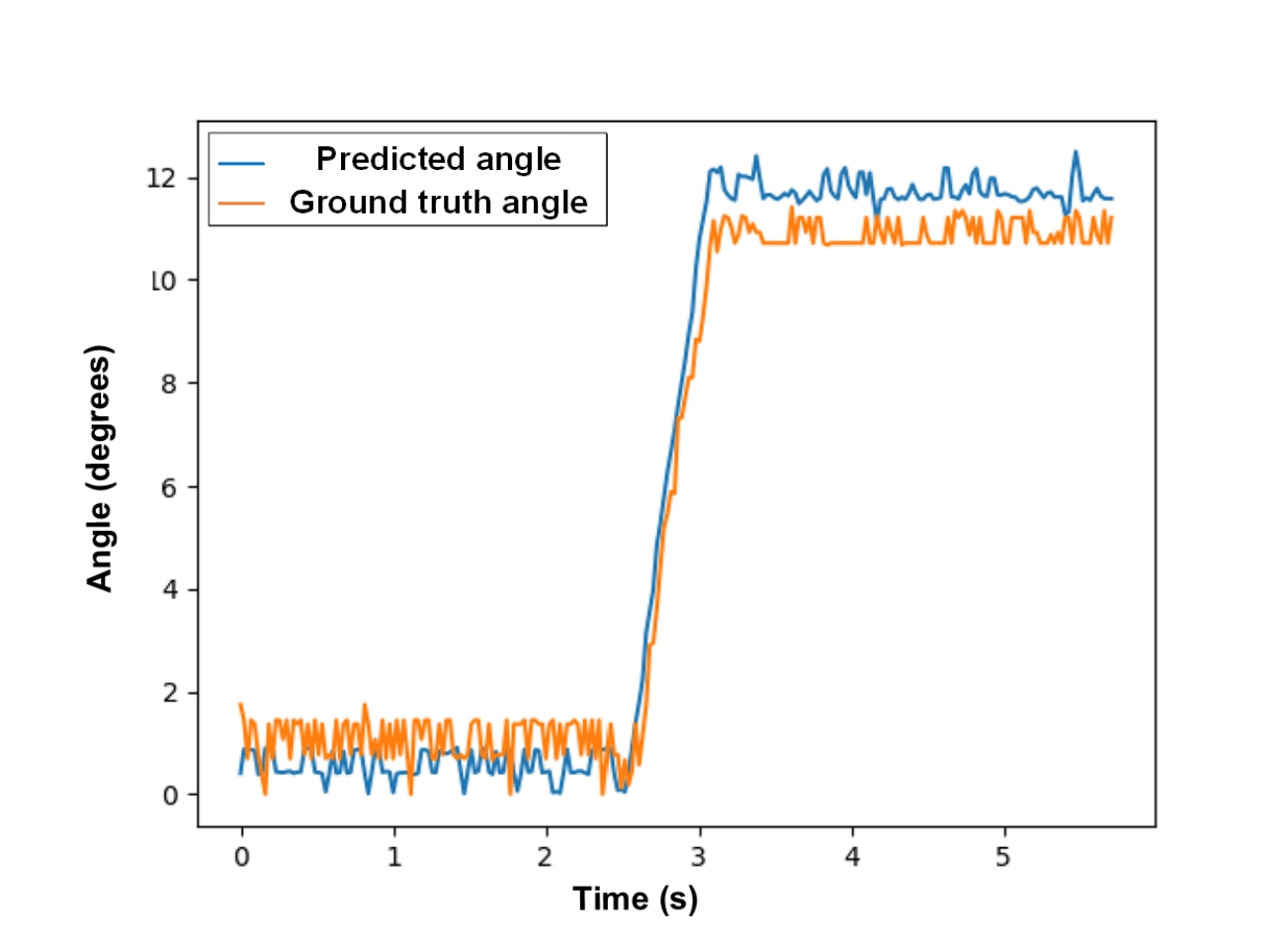}
     \end{subfigure}
    \begin{subfigure}[b]{0.24\textwidth}
         \centering
         \includegraphics[width=\textwidth, height=4cm]{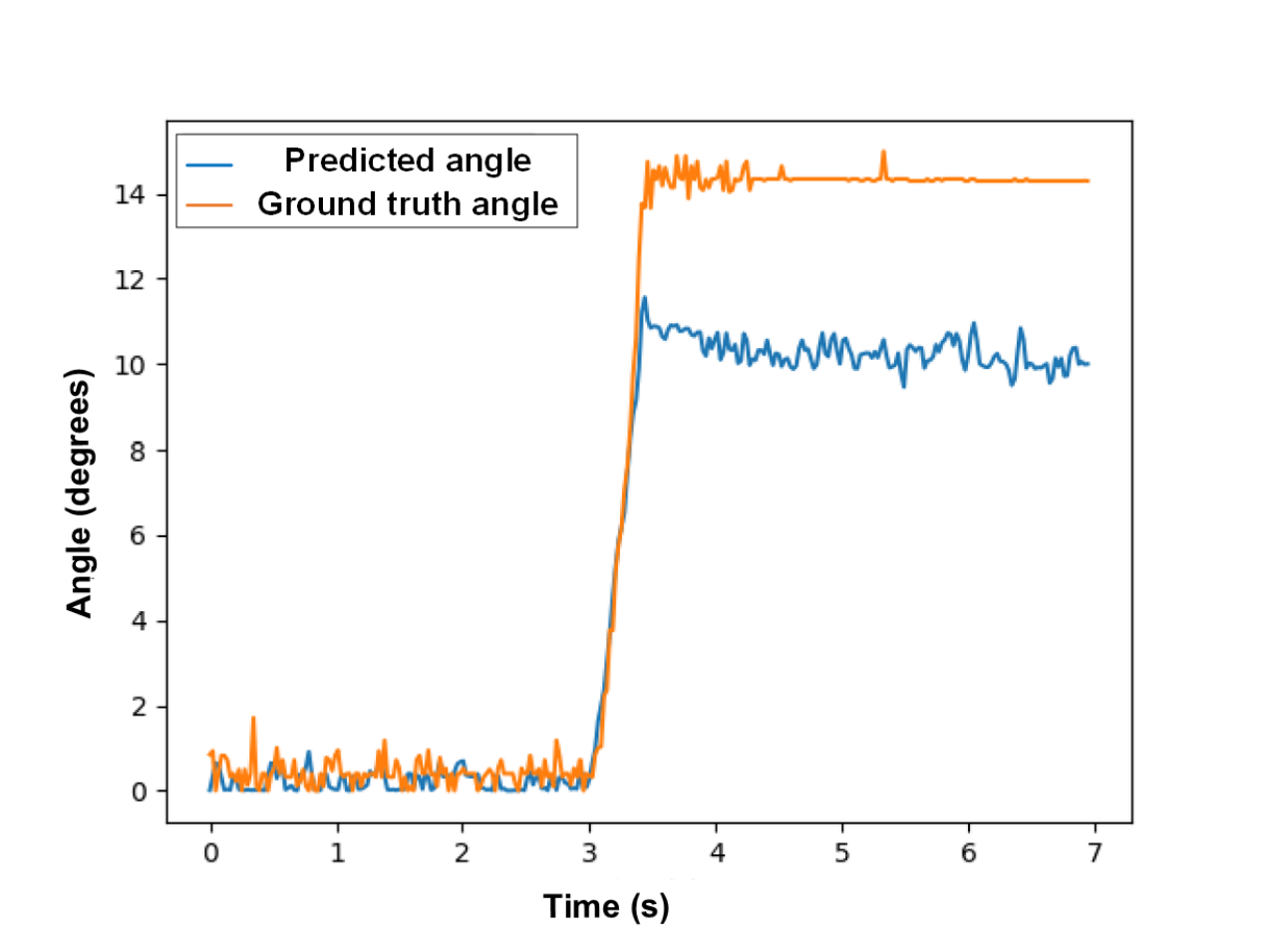}
     \end{subfigure}

     \begin{subfigure}[b]{0.24\textwidth}
         \centering
         \includegraphics[width=\textwidth, height=4cm]{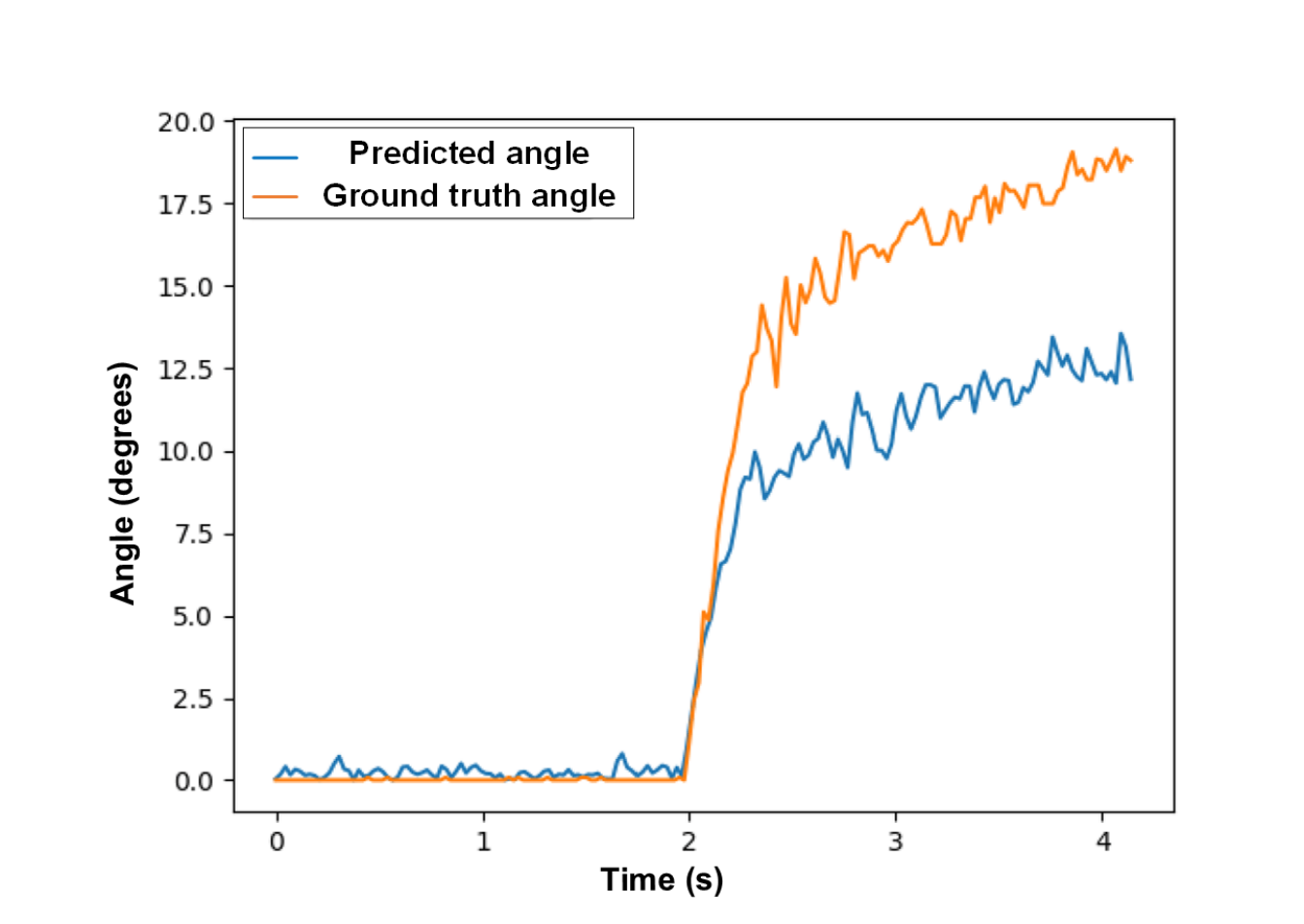}
     \end{subfigure}
     \begin{subfigure}[b]{0.24\textwidth}
         \centering
         \includegraphics[width=\textwidth, height=4cm]{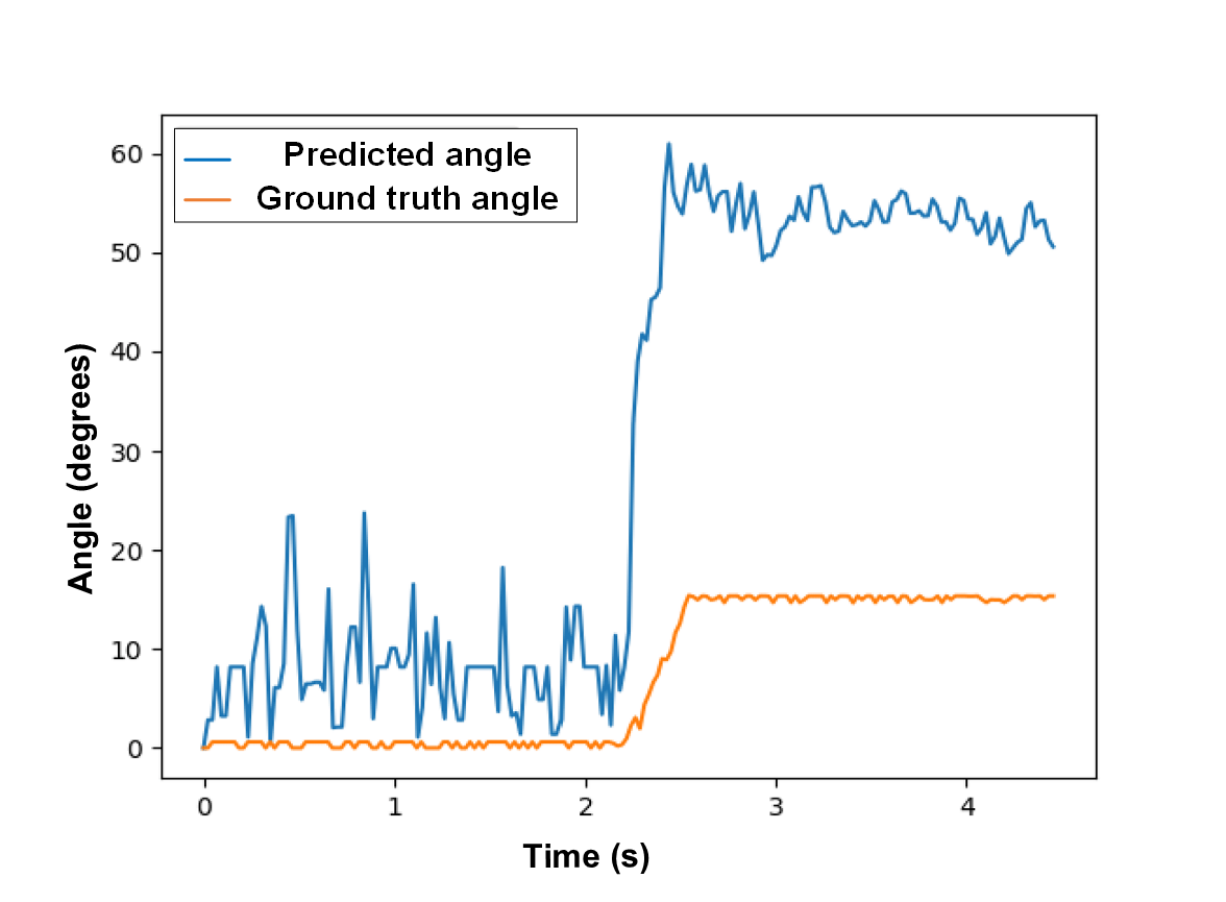}
     \end{subfigure}
     
      \caption{Angle calculation with different grasping surfaces. First row: edge surface, second row: cylindrical surface} 
     \label{fig:limitation2}
\end{figure}

Although this work proves that our approach achieves promising results, our system also has limitations. For example, when the segmented contact region is like a circle. In this case, our proposed method converts a circular mask into a single point rather than a line after applying the Skeleton transformation, as described in Fig. \ref{fig:limitations1}, and the predicted angle is not, therefore, reliable. This limitation could be solved by using a vision system that calculates the grasping points on a surface with a low curvature or edge. In Fig. \ref{fig:limitation2}, we show how the angle calculation varies depending on the grasping surface by using two different objects from Fig. \ref{fig:limitations1}. If the surface is like an edge, the results are promising, but the calculation of the angle is less reliable with surfaces with large curvature. Non-homogeneous objects which generate non-continuous contact shapes are also a limitation, although only closed contact shapes are considered in this work.

\section{CONCLUSIONS}
\label{sec:conclusions}
In this proposal, we propose a two-stage system with which to measure the rotation angle caused by slippage events 
when we work with DIGIT tactile sensors whose tactile reading is an RGB image. Our proposal uses TSNN based on PSPNet to obtain the local contact region between the robot’s fingertips and the manipulated object, after which the Skeleton Thinning algorithm estimates 
the rotation angle. Our method 
achieves a MARE of \textbf{1.85º$\pm$0.96º} when subjectively compared with the error of \textbf{3.96º $\pm$ UNK} in \cite{kolamuri} and the error of \textbf{4.39º $\pm$ 0.18º} in \cite{toskov}, and objectively compared with the error of \textbf{3.23º $\pm$ 1.69º} in \cite{comparison}. Moreover, the code is available at \href{https://github.com/AUROVA-LAB/aurova_grasping/tree/main/Tactile_sensing/Digit_sensor/Tactile_segmentation/code}{\textbf{github link}} and a demo is shown in the following \href{https://www.youtube.com/watch?v=AsnikqjCa_w}{\textbf{video link}}. Our system also has some limitations regarding the shape of the contact region. When this shape is nearly a circle, it is impossible to estimate its rotation movement. In this case, we propose to grasp the object by surfaces with a small curvature and show how
this improves the results in comparison with large curvature surfaces. On the other hand, objects such as a spiked ball, which generate more than one contact shape, would make the estimated angle less reliable. In addition, as future work, we are working on solving these two limitations to expand our method to detect slippage when grasping more complex objects.


\end{document}